\documentclass[lettersize,journal]{IEEEtran}
\usepackage{amsmath,amsfonts}
\usepackage{algpseudocode}
\usepackage{algorithm}
\usepackage{setspace}
\usepackage{array}
\usepackage[font=footnotesize,labelfont=sf,textfont=sf]{caption}
\usepackage{textcomp}
\usepackage{stfloats}
\usepackage{url}
\usepackage{verbatim}
\usepackage{graphicx}
\usepackage{cite}
\usepackage[font=footnotesize,labelfont=sf,textfont=sf]{subcaption}
\usepackage{tabularx}
\graphicspath{{figures/}}

\DeclareRobustCommand{\IEEEauthorrefmark}[1]{\smash{\textsuperscript{\footnotesize #1}}}

\begin{document}

\title{Spherical acquisition trajectories for X-ray computed tomography with a robotic sample holder}

\author{\IEEEauthorblockN{Erdal Pekel\IEEEauthorrefmark{1,2},
    Martin Dierolf\IEEEauthorrefmark{2,3},
    Franz Pfeiffer\IEEEauthorrefmark{2,3,4}, and
    Tobias Lasser\IEEEauthorrefmark{1,2}}\\
  \IEEEauthorblockA{\IEEEauthorrefmark{1}Department of Computer Science, School of Computation, Information and Technology, Technical University of Munich, Garching, Germany}\\
  \IEEEauthorblockA{\IEEEauthorrefmark{2}Munich Institute of Biomedical Engineering, Technical University of Munich, Garching, Germany}\\
  \IEEEauthorblockA{\IEEEauthorrefmark{3}Department of Physics, School of Natural Sciences, Technical University of Munich, Garching, Germany}
  \IEEEauthorblockA{\IEEEauthorrefmark{4}Department of Diagnostic and Interventional Radiology, School of Medicine, and Klinikum rechts der Isar, Technical University of Munich, Munich, Germany}}

\maketitle

\begin{abstract}
  This work presents methods for the seamless execution of arbitrary spherical trajectories with a seven-degree-of-freedom robotic arm as a sample holder.
  The sample holder is integrated into an existing X-ray computed tomography setup.
  We optimized the path planning and robot control algorithms for the seamless execution of spherical trajectories.
  A precision-manufactured sample holder part is attached to the robotic arm for the calibration procedure.
  Different designs of this part are tested and compared to each other for optimal coverage of trajectories and reconstruction image quality.

  We present experimental results with the robotic sample holder where a sample measurement on a spherical trajectory achieves improved reconstruction quality compared to a conventional circular trajectory.
  Our results demonstrate the superiority of the discussed system as it outperforms single-axis systems by reaching nearly 82\% of all possible rotations.

  The proposed system is a step towards higher image reconstruction quality in flexible X-ray CT systems.
  It will enable reduced scan times and radiation dose exposure with task-specific trajectories in the future, as it can capture information from various sample angles.
\end{abstract}

% \begin{IEEEkeywords}
%   X-ray computed tomography, robotic arm, sample holder, spherical trajectory, reachability
% \end{IEEEkeywords}

\section{Introduction}\label{sec:introduction}
In recent years, industrial robotic arms have become more affordable for a broader audience thanks to the mass production of electrical and mechanical components.
At the same time, their control and software integration has become more feasible with the emergence of open-source software initiatives that aim to standardize and simplify the use of such robotic arms.
The lower entry barrier for robotic arms also offers new opportunities for X-ray computed tomography systems which benefit significantly from the improved flexibility.
Capturing two-dimensional absorption images from arbitrary poses can increase reconstruction image quality and completeness in the short term.
In the long term, tomographic systems would benefit from task-specific trajectories that can be generated and executed more efficiently with a more flexible sample holder \cite{fischer2016object}.

In recent work, we introduced a flexible robotic arm with seven degrees of freedom (DoF) as a sample holder within a laboratory X-ray computed tomography (CT) setup \cite{Pekel_2022}.
The arm adds flexibility to the setup as a sample holder by enabling arbitrary rotations and placement of the sample. Hence, it allows non-standard trajectories, as opposed to conventional circular or helical trajectories, that are not restricted in their sequence.
We also introduced a suitable calibration mechanism to determine the exact positioning of the sample from the image, as the values reported by the sensors of the robotic arm are not precise enough for reconstruction purposes.
The calibration mechanism requires a sample holder part attached to the robotic arm, which was also introduced in \cite{Pekel_2022}.

In the following, we present our work on optimizing various aspects of the robotic sample holder for the seamless execution of spherical trajectories.
We modified the sample holder part attached to the robotic arm to improve coverage of spherical trajectories.
We provide a detailed analysis of the different shapes of this sample holder part and their effect on the execution of different types of trajectories.
We also present experimental results demonstrating the improved performance of spherical trajectories our system can execute.
We provide a quantitative comparison of reconstruction results to conventional trajectories.

In the remainder of this section, we will provide an overview of related work on imaging with robotic arms and trajectory optimization methods for imaging systems.

% \subsection{Related work on X-ray CT with Robotic arms}\label{sec:related-work-xray-ct-robotic-arms}
Robotic arms were also used in the past in computed tomography systems \cite{ziertmann2020robot, Landstorfer2019,siemens-artis-zeego}.
The main difference to our work is the kind of robotic arm used.
Due to its seven DoF, ours offers higher flexibility than the robotic arms used in related work.

In \cite{ziertmann2020robot}, the source and detector are mounted on robotic arms which move such that the sample is centered between the source and the detector.
The main differences to our work are that this is not a laboratory scale setup but an assembly line scale and that the sample does not move but the source and detector.
The authors do not specify the exact models of the robotic arms, but from the figure, we can see that they have five DoF (compared to seven with our robotic arm).
The reduced DoF count means the robot is less flexible and has difficulties reaching certain acquisition angles.

In \cite{Landstorfer2019}, the authors demonstrate the advantage of non-circular CT scanning trajectories.
The experiments are conducted in a simulation using a 3d model of the specimen.
With a circular scanning trajectory, the specimen completely absorbs X-rays in areas where highly-absorbing spheres are added, and reconstruction quality is impacted, resulting in streak artifacts.
The authors demonstrate by simulating a six DoF robotic arm that a non-circular trajectory that almost covers the entire rotational sphere would be possible and result in a reconstruction with no artifacts.

The Siemens Healthineers \textit{Artis pheno} angiography platform \cite{siemens-robotic-xray} consists of a single five DoF robotic arm that moves the detector and source with a fixed distance between each other.
The main differences to our work are that the detector and source are both mounted to the robotic arm and hence are moving parts.
The robotic arm is positioned such that the patient's body part, which is of interest, lies precisely between the source and the detector.
This system does not support arbitrary trajectories.

In \cite{twin-robotic-xray-herl}, the authors present an X-ray tomography system with two robotic arms.
The X-ray source and the detector can be moved independently from each other by the two arms, and the sample is mounted statically between them.
There are two key differences to the system that we propose.
The first is that we are moving the sample, and thus our system only requires one robotic arm instead of two.
Furthermore, moving the sample does not restrict the system to movable X-ray sources and detectors; hence, it is more flexible.
On the other hand, this means that we restrict the samples in our system in size and weight, and the sample should not deform or otherwise change when moved.
The second key difference is that our robotic arm is significantly smaller and thus fits into an existing X-ray CT setup. In contrast, the robotic arms used in \cite{twin-robotic-xray-herl} can reach up to 3 meters of height (compared to 1.2 meters with our arm), which might not even fit into an existing laboratory \cite{kuka-kr-specs,panda-specs}.
Furthermore, the smaller robotic arm in our system is more affordable.
The most significant technical differences are that our robot has an additional degree of freedom (ours seven vs. theirs six) but worse repeatability when the identical trajectory is executed repeatedly (ours 0.1 mm vs. 0.04 mm theirs).
The higher degree number increases the probability of our system reaching positions on complex trajectories.
In contrast, our system tackles its more inaccurate repeatability with the calibration procedure described in \cite{Pekel_2022}.

Our sample holder is designed and optimized for executing task-specific trajectories.
The authors of \cite{twin-robotic-xray-herl} use their system to present a trajectory optimization approach that optimizes image quality.
% The authors demonstrate that an optimized task-specific trajectory can be computed based on a predefined set of measurements called base scans.
% The resulting trajectory can reduce scan times and radiation dose while maintaining reconstruction image quality.
% In the presence of metal parts, increasing image quality for the reconstructions is demonstrated only with simulated measurements and for an existing dataset with a non-circular trajectory where they calibrated the sample position manually.
% Our robotic sample holder is well suited for executing the base scans and, subsequently, the optimized trajectories proposed in \cite{twin-robotic-xray-herl}.
Existing work on trajectory optimization relies heavily on simulations of X-ray projections with models of known samples (see \cite{Hatamikia2022}) because of a lack of sample holders that can place the sample at arbitrary rotations.
When employing robotic systems that tackle this limitation, the placement accuracy of the robotic arm is not high enough for CT purposes.
For the reconstruction of the measurements, we need a suitable calibration mechanism that extracts the geometry of the sample \cite{Pekel_2022}.
However, the systems mentioned above lack this feature.
Our system solves this issue and hence can execute arbitrary trajectory types in practice.

In this work, we will outline our approach toward executing task-specific trajectories by investigating the challenges of executing non-standard trajectories for spherical trajectories.
We will conclude the feasibility of using a robotic arm as a sample holder for arbitrarily complex trajectories and the limitations we must tackle.

\section{Methods}\label{sec:methods}
In this section, we discuss the methods for executing spherical trajectories with the robotic arm as a sample holder in a laboratory X-ray CT setup.
After introducing the system's hardware components, we describe more specific aspects like path planning, sphere sampling, and reconstruction.
We also refer interested readers to \cite{Pekel_2022} for a more detailed introduction to the robotic sample holder for X-ray CT.

\subsection{Hardware setup}
\begin{figure*}[t]
  \centering
  \subfloat[Lab setup]{\includegraphics[width=.52\textwidth]{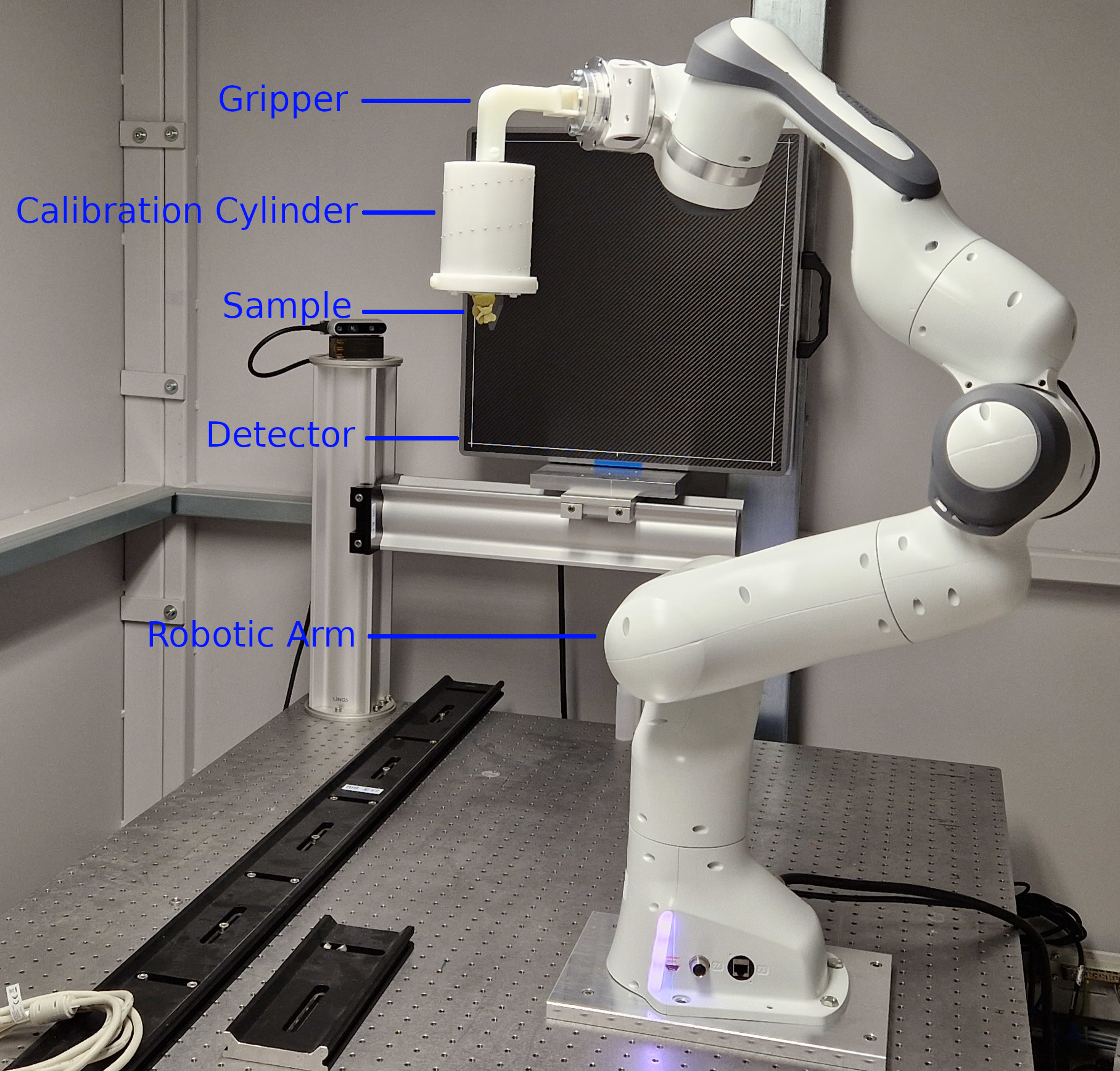}%
    \label{fig:lab_photo}
  }
  \hfil
  \subfloat[Straight gripper]{\includegraphics[width=.27\textwidth]{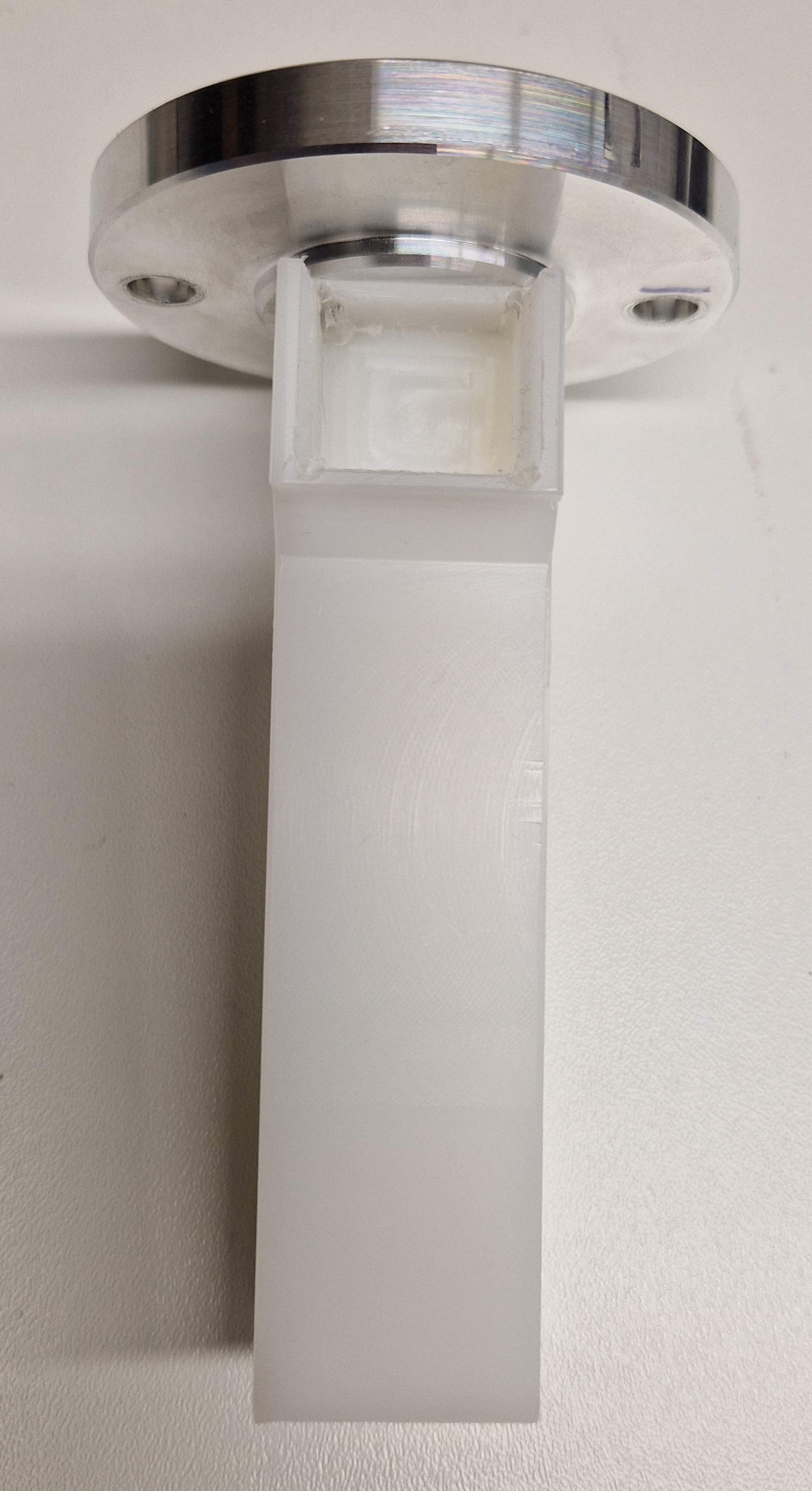}%
    \label{fig:straight-gripper}
  }

  \medskip

  \subfloat[Cylinder part with calibration structure (helix)]{\includegraphics[width=.55\textwidth]{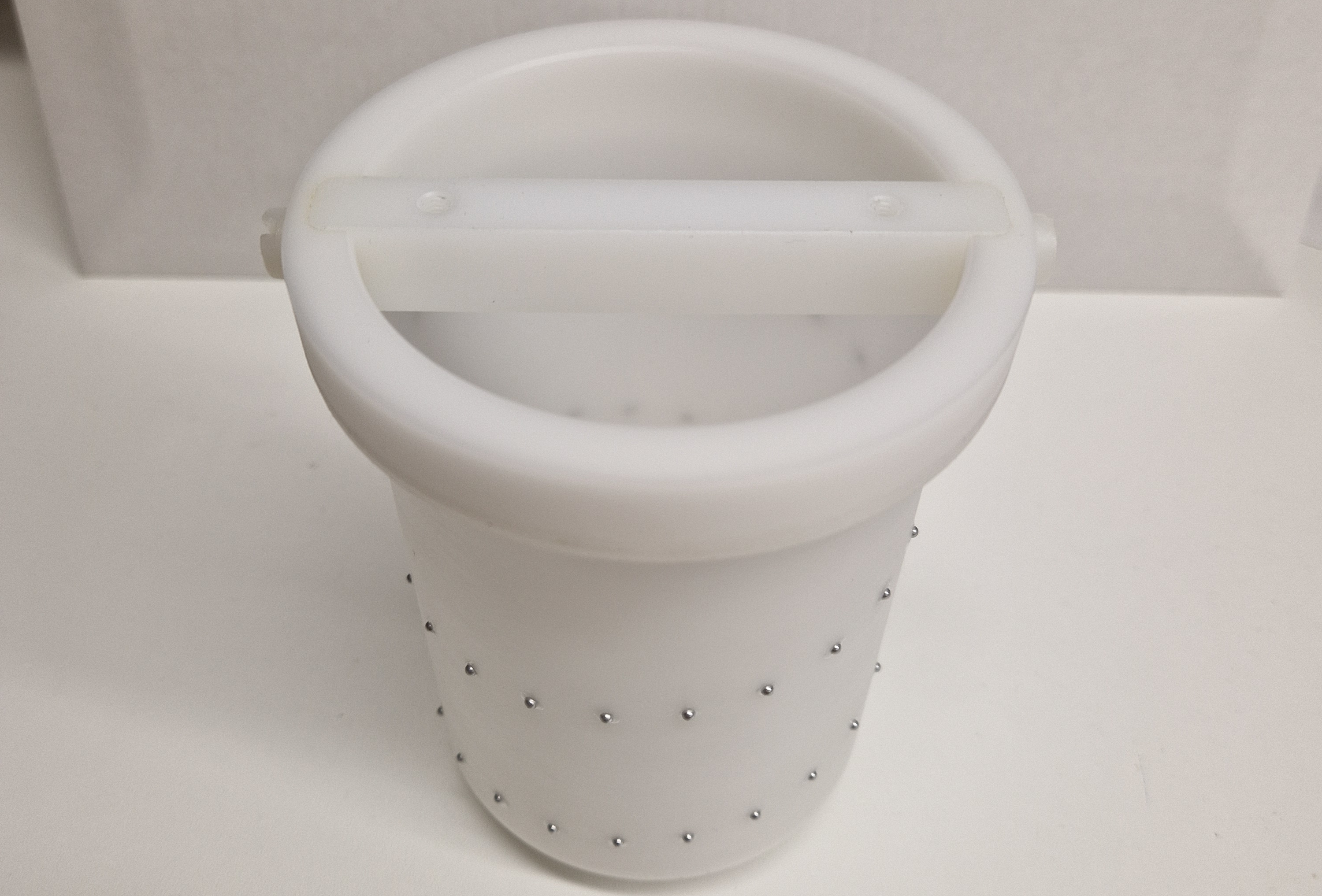}%
    \label{fig:sample-holder-cylinder}
  }
  \hfil
  \subfloat[Curved gripper]{\includegraphics[width=.35\textwidth]{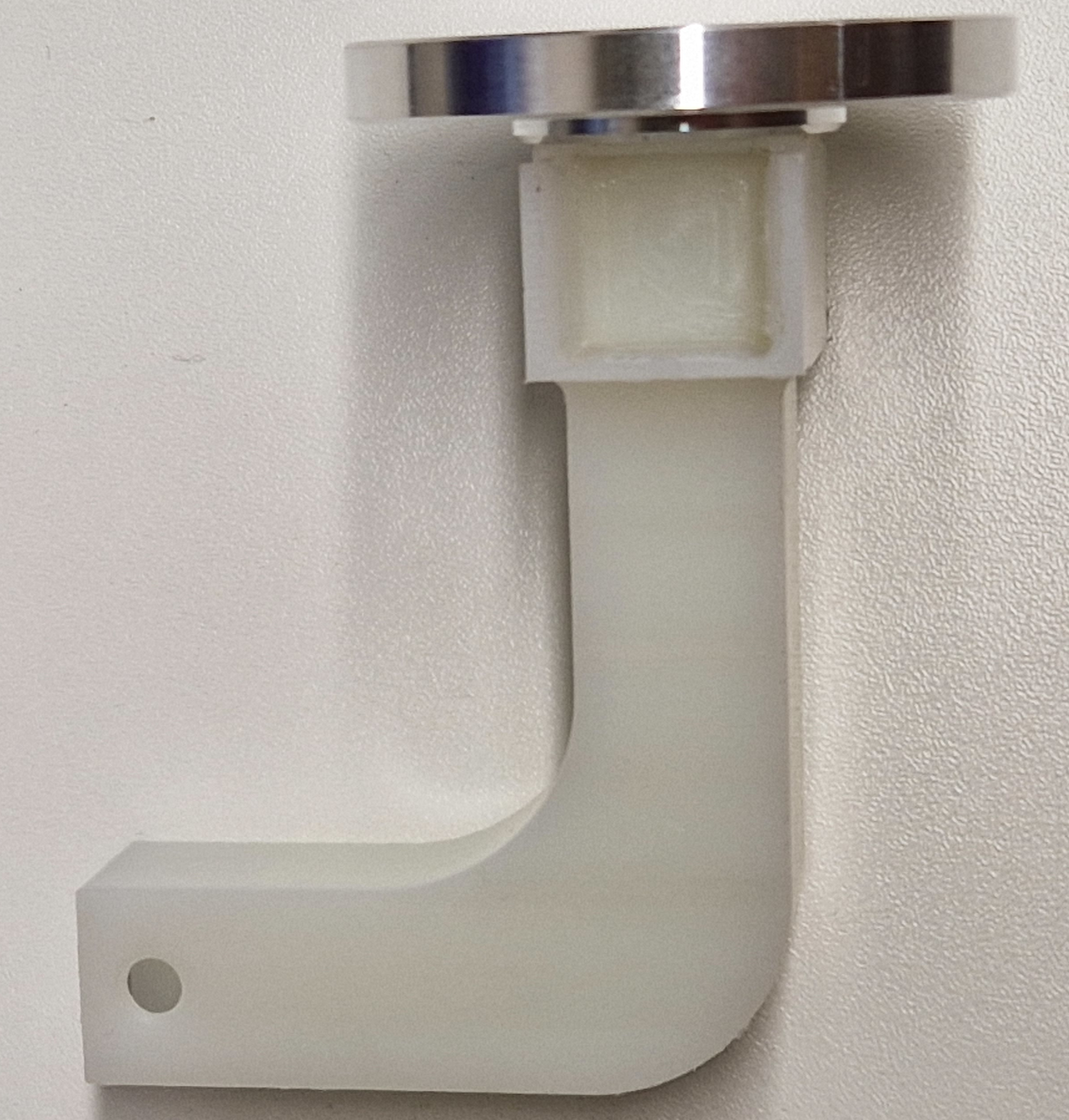}%
    \label{fig:curved-gripper}
  }

  \caption{
    \textbf{Hardware setup}.
    In (\subref{fig:lab_photo}), the robotic arm is mounted on a table with the source and the detector inside a safety hutch.
    The source-to-robot distance is 136 cm, and the robot-to-detector distance is 79 cm.
    Two depth cameras monitor the robot's movement and stop the robot controller when the executed trajectory interferes with obstacles.
    A power switch can also stop the robotic arm. It routes to the operator's table outside of the hutch.
    In (\subref{fig:straight-gripper}), (\subref{fig:sample-holder-cylinder}), and (\subref{fig:curved-gripper}), the different parts of the sample holder part are displayed.
    (\subref{fig:straight-gripper}) and (\subref{fig:curved-gripper}) depict the two gripper types, and (\subref{fig:sample-holder-cylinder}) depicts the cylinder part, which houses the geometric structure for calibration.
    The gripper part is mounted to the cylinder part for experiments.
  }
  \label{fig:hardware_setup}
\end{figure*}

Fig. \ref{fig:hardware_setup} displays the system's hardware components.
The main difference to a conventional X-ray CT setup is the seven DoF robotic arm \textit{Panda} from the manufacturer FRANKA EMIKA \cite{panda-specs}.
It has a maximum reach of $855$ mm and a repeatability of $0.1$ mm when repeatedly moved from a specific starting position to a goal position on a fixed trajectory.
It has two fingers that can move on a linear axis and grasp objects.
The maximum allowed payload is $3$ kg.

Two \textit{Intel Realsense D435} depth cameras capture the robot's movements and provide 3d information about the surroundings as a point cloud.
The cameras are connected directly to the workstation with the control software for use by the collision detection mechanism described in \cite{Pekel_2022}.

The robotic arm is mounted on an optical table inside a radiation shielding enclosure for X-ray CT, which houses the X-ray source and the detector (see Fig. \ref{fig:lab_photo}).
The detector (Varex XRD 4343) has a maximum resolution of $2880 \text{x} 2880$ and connects to a different workstation on the network that exposes the raw 16-bit grayscale images.

A power switch can turn the robotic arm off in an emergency from outside the safety hutch.

\subsection{Sample holder part}\label{sec:sample_holder}
The sample holder part is a critical component of the system as it allows the robotic arm to grasp samples of arbitrary shape and is a fundamental part of the calibration process that identifies the position and orientation of the sample.
The sample holder part and the gripper components are visualized in Figures \ref{fig:straight-gripper}, \ref{fig:sample-holder-cylinder}, and \ref{fig:curved-gripper}.
It consists of two parts.
The gripper part is where the robot's fingers can grasp the holder steadily.
The cylinder part fulfills the purpose of placing a helix of fiducial markers on a cylinder next to the sample.
The lower part is called \textit{gripper part} throughout this paper, and it mounts directly to the last link of the robotic arm after unmounting the hand from the arm.
We will discuss reasons for using the arm without the hand in section \ref{sec:path_planning}.
Before attaching the sample holder part to the robotic arm, the sample needs to be glued to the mounting plate, which is inserted into the cylinder from the top at the intended position (see Fig. \ref{fig:sample-holder-cylinder}).

The cylinder is $118$ mm tall and $50$ mm in diameter inside.
The sample holder part was designed with the 3d modeling software \textit{Autodesk Inventor} and manufactured by a cutting machine from a solid piece of polyoxymethylene (POM).

The reference structure embedded in the sample holder is a helix comprising 50 embedded aluminum spheres of 2 mm diameter.
These spheres were fixed manually in notches included in the holder's design process.
The spheres appear as circles on the detector images we segment during calibration.

Compared to our previous work\cite{Pekel_2022}, we modified the sample holder part by introducing a gripper part that connects the robotic arm's last link with the holder's cylindrical part.
This new part enables the use of different gripper shapes (straight, curved) in varying lengths, which enables the robotic arm to reach different areas of more specific trajectories, for example, spherical trajectories.
We mount the gripper part to the cylinder with a screwing mechanism.

\subsection{Path planning and robot control}\label{sec:path_planning}
With the path planning procedure, our system exposes an abstract interface to the user for planning and executing advanced trajectories.
A detailed description of the path planning process can be found in our previous publication \cite{Pekel_2022}.

A trajectory comprises a series of \textit{way-points} that the robotic arm approaches in the given order.
The arm stops at each way-point, triggering the detector to capture an image.
After successful image acquisition, the robotic arm continues trajectory execution.
The user specifies the parameters for sampling the way-points on the trajectory from the user interface.
Given these parameters, the system first samples the way-points depending on the trajectory type.
Subsequently, the underlying motion planning pipeline plans a path from each way-point to its successor.
If the arm cannot reach way-point $i+1$ from way-point $i$, e.g., because there is no collision-free path, we plan a path from $i$ to way-point $i+2$, and $i+1$ is marked as not reachable.
The unsuccessful planning means that we know that no detector image will be captured later for way-point $i+1$ before executing the trajectory.
Finally, we connect these paths, and the output is a trajectory that starts at the robotic arm's current position and passes all way-points in the given order.

Problems can still arise while executing the successfully planned trajectory.
The motion planning pipeline plans the paths between the way-points based on an internal model that the manufacturer provides.
This model includes the kinematic and collision model of the robotic arm.

The kinematic model defines the arm's link lengths, joint types, and joint limits.
The manipulation software constructs a \textit{configuration space} from this model with $n$-dimensions, where $n$ is the number of joints.
The software samples this space at a fixed density and connects the sampled points with an exploration algorithm.
The result is a graph where each vertex represents a collision-free and valid configuration for the robotic arm.

The collision model marks specific areas of the \textit{configuration space} as colliding configurations of the arm, which protects the arm from colliding with itself in specific configurations.
These collisions are possible because driving the joints within their valid ranges can easily result in a configuration where certain links collide with each other; for example, the last link can easily reach the first link.
The surroundings of the robotic arm are also added to the configuration space (e.g., the detector and the table) as collision objects, invalidating possible configurations of the robotic arm in which it would collide with these objects.
We discussed this in our previous work under \textit{passive collision detection} \cite{Pekel_2022}.

Incorporating these two models into our path-planning process should be enough to plan safe trajectories.
In practice, issues arise because the controller box only receives part of the planned trajectory in advance from beginning to end.
It sequentially receives joint commands of the current trajectory one by one.
As a result, there is uncertainty in the control box about the upcoming commands for the joints, as they might trigger a collision.
For this reason, the control box that receives the joint commands one by one interpolates the upcoming command by combining the latest commands and the current state to check if the arm's links will collide in the next instant.
If this check is positive, an internal collision avoidance reflex is triggered, and the control box stops the robotic arm.
We trigger the error recovery routine provided by the manufacturer in order to recover from this state.
The links are moved slightly out of our planned trajectory for this purpose.
We have increased the sizes of the robotic arm's links in the internal collision model of the robotic arm in order to reduce the number of these reflexes.

The error recovery routine mentioned above leads to an undefined state in our planned trajectory because the resulting position of the arm is unexpected, as it lies outside of the planned trajectory.
We have implemented an error recovery routine for our trajectory that will plan to a fixed intermediate way-point (\textit{homing state}) and set a flag to ensure we do not use the image recorded there in later steps of our pipeline.
Afterward, we plan from the homing state to the next scheduled way-point, where the handover to the initially planned trajectory will occur.

Our software package calculates a unique identifier for the current state of the robotic arm's environment, the \textit{configuration space}.
It also calculates a unique identifier for the trajectory that the user requested.
Both identifiers are calculated with a \textit{sha256} hash of the requested parameters (trajectory) and contained elements (configuration space).
This calculation is an additional safety measure as it maps different compositions of the robotic arm's environment to a unique identifier.
We can save a planned trajectory to the file system for repeated use in the same environment with a unique ID for the trajectory.
When the user requests a trajectory with the same parameters as before, we check for the file on disk and send the trajectory directly to the robotic arm for execution.
Additionally, the trajectory ID ensures that in combination with the environment ID, we only execute a trajectory from the filesystem when it suits the current environment, including the surroundings of the robotic arm, the sample holder, and the attached sample.

\subsection{Sphere sampling}\label{sec:sphere_sampling}
In order to generate a spherical trajectory, we need to sample points covering the surface of a sphere.
Each point represents a rotation of the sample.

Sampling a fixed number of $n$ points on the sphere is a well-studied problem \cite{Saff1997, Rafaely2019, Khalid2014, Hardin1995}.
The goal is to distribute the points uniformly on the sphere's surface.
Trivial approaches like sampling on the two polar axes independently and combining the samples to get 3d coordinates do not lead to a uniform sampling on the sphere surface.
In this work, we utilized the Hierarchical Equal Area isoLatitude Pixelization (\textit{HEALPix}) for this purpose \cite{Gorski2005}.

HEALPix partitions the sphere's surface into a fixed number of areas of equal size.
The centers of these areas are the sampled points on the sphere.
The discretization number $N_{pix}$ determines the number of points on the sphere.
In HEALPix, the grid resolution parameter $N_{side}$ determines the number of pixels and hence the number of points sampled on the sphere: $N_{pix} = 12 \cdot N_{side}^{2}$, where $N_{side} \in \mathbb{N}$.
The user chooses $N_{side}$, and $N_{pix}$ is calculated from that parameter.

If the user needs a specific number of points on the sphere, we can choose a higher grid resolution parameter $N_{side}$ in the first step, and a smaller number of points can be sampled from the resulting grid in a later step.
For our experiments, this did not present a problem as we chose $N_{side} = 10$, which results in $N_{pix} = 12 \cdot 10^{2} = 1200$ pixels on the grid and hence $1200$ potential way-points on the spherical trajectory.% assuming a collision-free path is available (see section \ref{sec:path_planning}).

\subsection{Calibration}\label{sec:calibration}
The calibration procedure tackles the issue that the robotic arm does not sufficiently accurately place the sample at the desired position due to inaccurate path planning and inaccurate electrical motors at its joints.
Reading the sensors of the robotic arm and deducing the sample's current position is also insufficient to determine the correct position, as the sensors report inaccurate values.
However, the reconstruction step requires the exact position of the sample at each view.
With the calibration procedure, we can identify the actual positions and orientations of the sample from the detector images.
For the calibration, a sample holder part with an embedded geometric reference structure we can identify on the detector images was necessary and introduced in section \ref{sec:sample_holder}.

The sample holder houses a helix structure with known parameters.
A set of points was sampled on this continuous structure to place calibration spheres at known locations (drilled holes on the cylinder part), which we will later segment on the detector images.
We determine the expected location of the helix structure on the detector images with a priori information about the system: the camera matrix, the robotic arm's location in the setup, and the values reported by the sensors at the arm's joints.
We can project the continuous helix structure onto the image and use a suitable cost function to determine the error between our guess for the helix structure and its actual position on the image by comparing it to the segmented circles.
We optimize the parameters for the translation and rotation of the geometric structure and the sample on a given image with the Levenberg-Marquardt non-linear optimization algorithm.
The sensor values from the robotic arm serve as an initial guess for the optimization.
A more detailed description of the calibration procedure is available in \cite{Pekel_2022}.

We have improved the calibration procedure by checking the segmented circles for false positives.
While testing the system with different samples in our lab, we experienced that the segmentation algorithm detected circles falsely on the screw heads and the samples.
We solved this issue by checking the distance (L2-Norm on 2d image) between the segmented circle centers and our guess for each calibration circle.
Suppose the distance of the given segmented circle's center is below a certain threshold for one of the guessed calibration circles.
In that case, we interrupt the checks and assume that the segmented circle is a calibration circle.
Suppose the calculated distance is above the fixed threshold for all calibration circles.
In that case, we assume that this specific circle is a false positive, which we do not consider in the later steps of the calibration procedure.
For example, for the experiment with the curved sample holder part on the spherical trajectory, identifying the false positives improved the number of successfully calibrated images from 912 to 946 out of 965.

\subsection{Reconstruction}\label{sec:reconstruction}
We used around $900$ equidistant X-ray projections for the tomographic reconstruction along a circular or spherical trajectory sized $720 \times 720$ pixels with a spacing of $600 \: \mu m$.
For the spherical trajectory, the number of projections varied by $\pm8$\% (see Table \ref{tab:reachability-statistics}).
The reconstruction volume consisted of $720 \times 720\times 720$ isotropic voxels with a spacing of $38 \: \mu m$.
We used our C++ reconstruction framework \textit{elsa} \cite{LasserElsa2019} to perform the reconstruction using an iterative conjugate gradient solver run for $30$ iterations on a Tikhonov regularized weighted least squares problem, with the Josephs method for X-ray transform discretization and cone beam geometry.
Further iterations showed no improvement in the cost function.

\subsection{Software stack}\label{sec:software_stack}
The central part of our software stack is the \textit{Robot Operating System} (\textit{ROS}) \cite{quigley2009ros}, which is a middleware for the communication of independent processes across a network.
We accomplish robot manipulation with the \textit{MoveIt!} framework \cite{coleman2014reducing,moveit-web} and the \textit{franka\_ros} configuration package \cite{frankaros}.
For image processing tasks and the circle segmentation, we use \textit{OpenCV} \cite{opencv_library}, for multithreading on the CPU \textit{OpenMP} \cite{Dagum1998OpenMPAI} and on the GPU \textit{OpenCL} \cite{5457293} and for the tomographic reconstruction \textit{elsa} \cite{LasserElsa2019}.
The scientific calculations in section \ref{sec:calibration} are implemented with \textit{scipy} \cite{2020SciPy-NMeth}.
We read the 3d mesh files of the robotic arm with OpenMesh \cite{botsch2002openmesh}.
The sphere discretization in section \ref{sec:sphere_sampling} was implemented with the HEALPix C++ and Python interfaces \cite{Gorski2005, Zonca2019}.
For calculating sha256 sums of the configuration spaces mentioned in section \ref{sec:path_planning}, we use crypto++ \cite{cryptopp}.
The 2D reachability maps in section \ref{sec:experiments-reachability-analysis} were generated using matplotlib, and basemap \cite{Hunter:2007}.

\section{Experiments and results}\label{sec:experiments-and-results}
When running the main experiments, we designed two different gripper parts for the sample holder part and analyzed their coverage (\textit{reachability}) for different trajectory types with the robotic arm.
Additionally, we modified the mounting mechanism of the robotic arm by removing the hand and gripper from the arm and mounting the sample holder part directly to the last link.
The cylindrical sample holder part from section \ref{sec:sample_holder} was used for all experiments for geometric calibration.
The collision detection algorithm was running in the background throughout these experiments.

\subsection{Reachability analysis}\label{sec:experiments-reachability-analysis}
The term \textit{reachability} describes the percentage of target way-points that the robotic arm can reach from all target way-points.
For example, for a spherical trajectory with a fixed number of uniformly sampled points on the sphere, the reachability states the percentage of points that the robotic arm can \textit{reach}, meaning that the motion planning pipeline could find a valid and collision-free configuration for the arm for that specific point and a path to that configuration.
% The path planning process is described in detail in section \ref{sec:path_planning}.

We have conducted experiments with two types of trajectories: circular and spherical.
We have used two different grippers for each trajectory type, resulting in four experiments with an identical sample.
Our goal with these experiments was to analyze the impact of the gripper types on the reachability of the two trajectory types.

\subsubsection{Sample holder gripper type}\label{sec:experiments-sample-holder-gripper-types}
We manufactured and used two different types of gripper parts for the sample holder part mentioned in section \ref{sec:sample_holder}: straight and curved (l-shaped, $90^{\circ}$).
Figures \ref{fig:straight-gripper} and \ref{fig:curved-gripper} show the grippers.
The straight gripper differs from the sample holder part used in our previous publication because it introduces an additional distance between the grippers and the cylindrical part of the sample holder.
When we neglect the gripping cave, the length of the straight part amounts to 70 mm.
We expected improved reachability of spherical trajectories by introducing extra distance between the robotic arm's last link and the sample mounted on top of the cylindrical part and positioned at the central X-ray.
The extra distance would increase the distance between the individual links of the arm when approaching a goal for specific configurations, which helps avoid collision reflexes of the arm.

\subsubsection{Trajectory type}\label{sec:experiments-trajectory-types}
We executed our experiments on circular and spherical trajectories.
We aimed to compare the proposed robotic system's performance on these trajectory types for reachability.
For the spherical trajectories, the sphere discretization parameter $N_{side}$ was set to 10, which resulted in 1200 points on the sphere.
For the circular trajectories, we sampled 900 points on the circle in order to match the number of reachable points on the spherical trajectory for better comparability of our results, as the robotic arm can reach almost all way-points on the circular trajectory compared to the spherical trajectory where the robotic arm cannot reach approximately a quarter of the way-points.

We have created detailed statistics on the failure rates of all experiments during each stage of our pipeline.
% The main objective was to analyze the reachability of the robotic sample holder for each experiment.
The most common issue is that the motion planning pipeline can not find a valid configuration for a way-point on the trajectory.
Other issues that can reduce the number of usable images for the reconstruction are collision reflexes triggered by the robotic arm (section \ref{sec:path_planning}) and erroneous calibration (section \ref{sec:calibration}).

We have created tables with detailed statistics on the three types of errors.
The number of reached way-points compared to the potential number of way-points are listed in table \ref{tab:reachability-statistics}.
The number of way-points the arm could reach without triggering reflexes during execution is listed in table \ref{tab:trajectory-execution-statistics}.
The number of acquired images for each successfully calibrated experiment is listed in table \ref{tab:calibration-statistics}.
For example, for the measurement with the straight gripper part and spherical trajectory, the system acquired and calibrated 860 images out of 1200 potential way-points.

\begin{table}
  \centering
  \caption{Trajectory poses statistics}

  \begin{tabular}{ c | c c }
    \hline
             & circular            & spherical            \\
    \hline
    straight & 900 / 900 (100 \%)  & 909 / 1200 (75.6 \%) \\
    curved   & 835 / 900 (92.8 \%) & 979 / 1200 (81.6 \%)
  \end{tabular}
  \subcaption{
    \textbf{Reachability statistics}: Way-points with successful motion planning / potential way-points.
  }
  \label{tab:reachability-statistics}

  \bigskip
  \centering
  \begin{tabular}{ c | c c }
    \hline
             & circular             & spherical           \\
    \hline
    straight & 900 / 900 (100.0 \%) & 867 / 909 (95.4 \%) \\
    curved   & 835 / 835 (100.0 \%) & 965 / 979 (98.6 \%)
  \end{tabular}
  \subcaption{
    \textbf{Trajectory execution statistics}: Successful execution by robotic arm / way-points with successful motion planning.
  }
  \label{tab:trajectory-execution-statistics}

  \bigskip
  \centering
  \begin{tabular}{ c | c c }
    \hline
             & circular            & spherical           \\
    \hline
    straight & 896 / 900 (99.6 \%) & 860 / 867 (99.2 \%) \\
    curved   & 833 / 835 (99.8 \%) & 946 / 965 (98.0 \%)
  \end{tabular}
  \subcaption{
    \textbf{Calibration statistics}. Successful calibration / successful execution by the robotic arm.
  }
  \label{tab:calibration-statistics}

\end{table}

\begin{figure*}[t]
  \centering
  \subfloat[Straight gripper - circular trajectory]{\includegraphics[width=.45\textwidth]{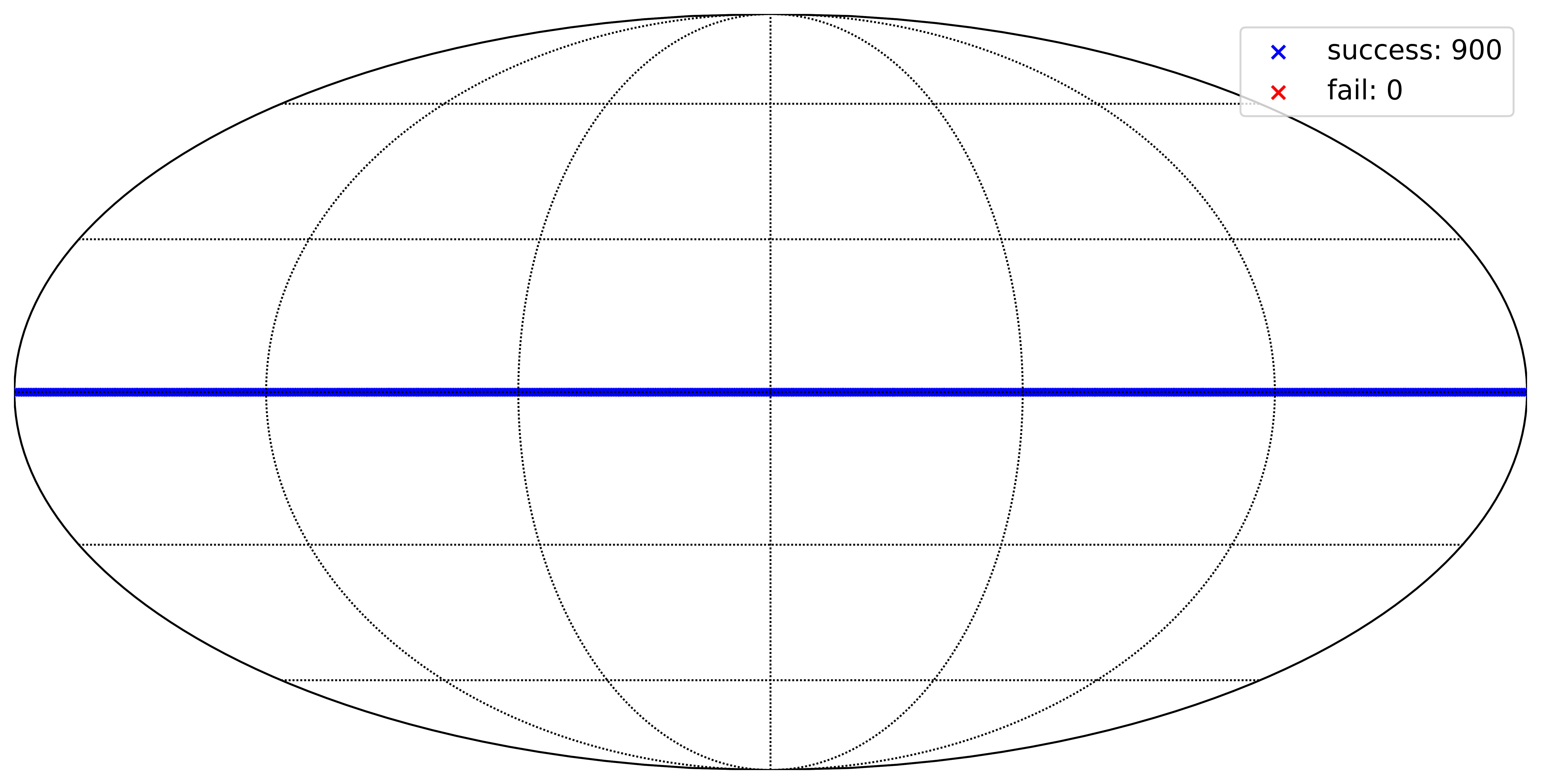}%
    \label{fig:trajectory-2d-straight-circular-direct}
  }
  \hfil
  \subfloat[Straight gripper - spherical trajectory]{\includegraphics[width=.45\textwidth]{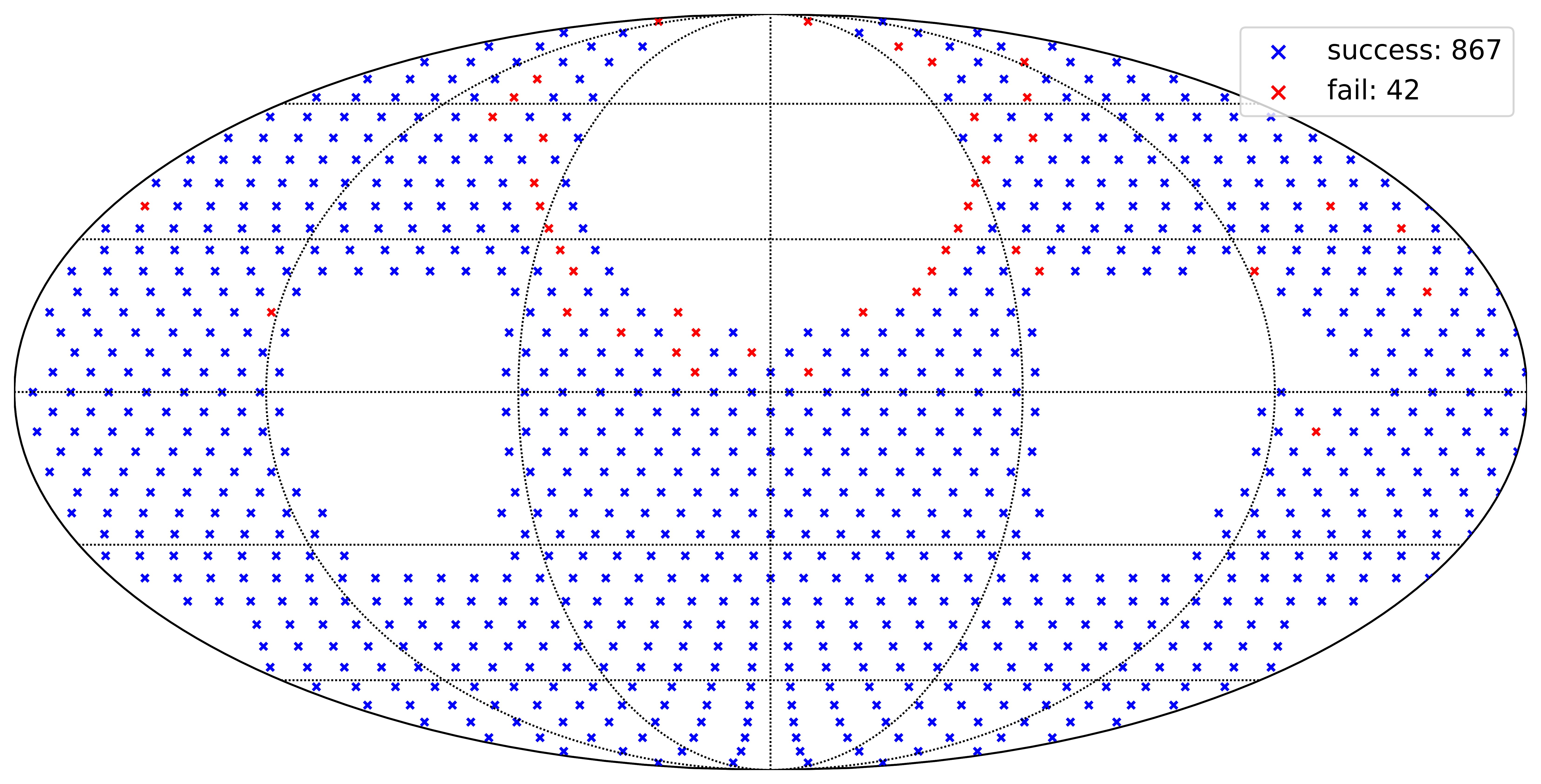}%
    \label{fig:trajectory-2d-straight-spherical-direct}
  }

  \medskip

  \subfloat[Curved gripper - circular trajectory]{\includegraphics[width=.45\textwidth]{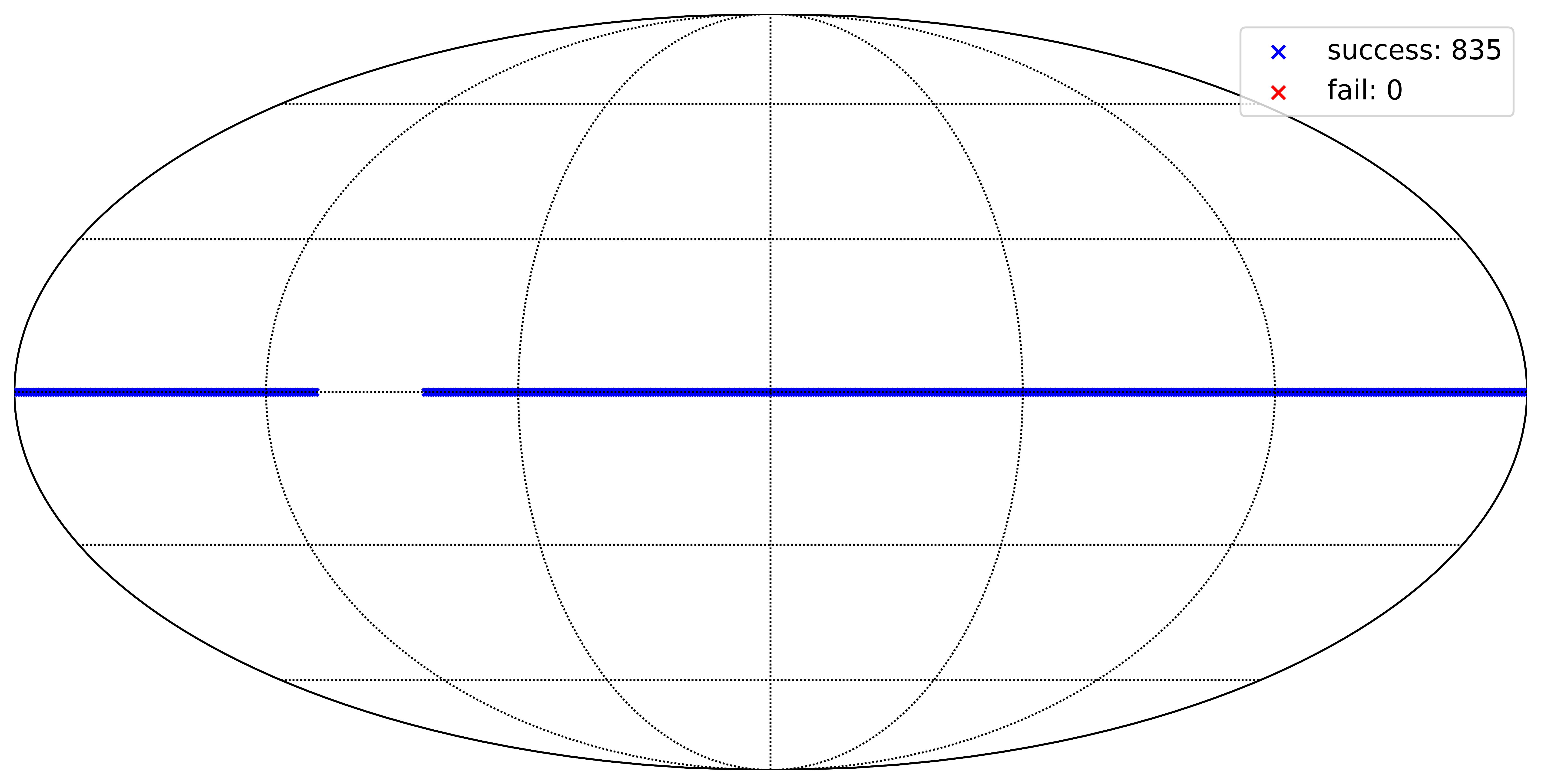}%
    \label{fig:trajectory-2d-curved-circular-direct}
  }
  \hfil
  \subfloat[Curved gripper - spherical trajectory]{\includegraphics[width=.45\textwidth]{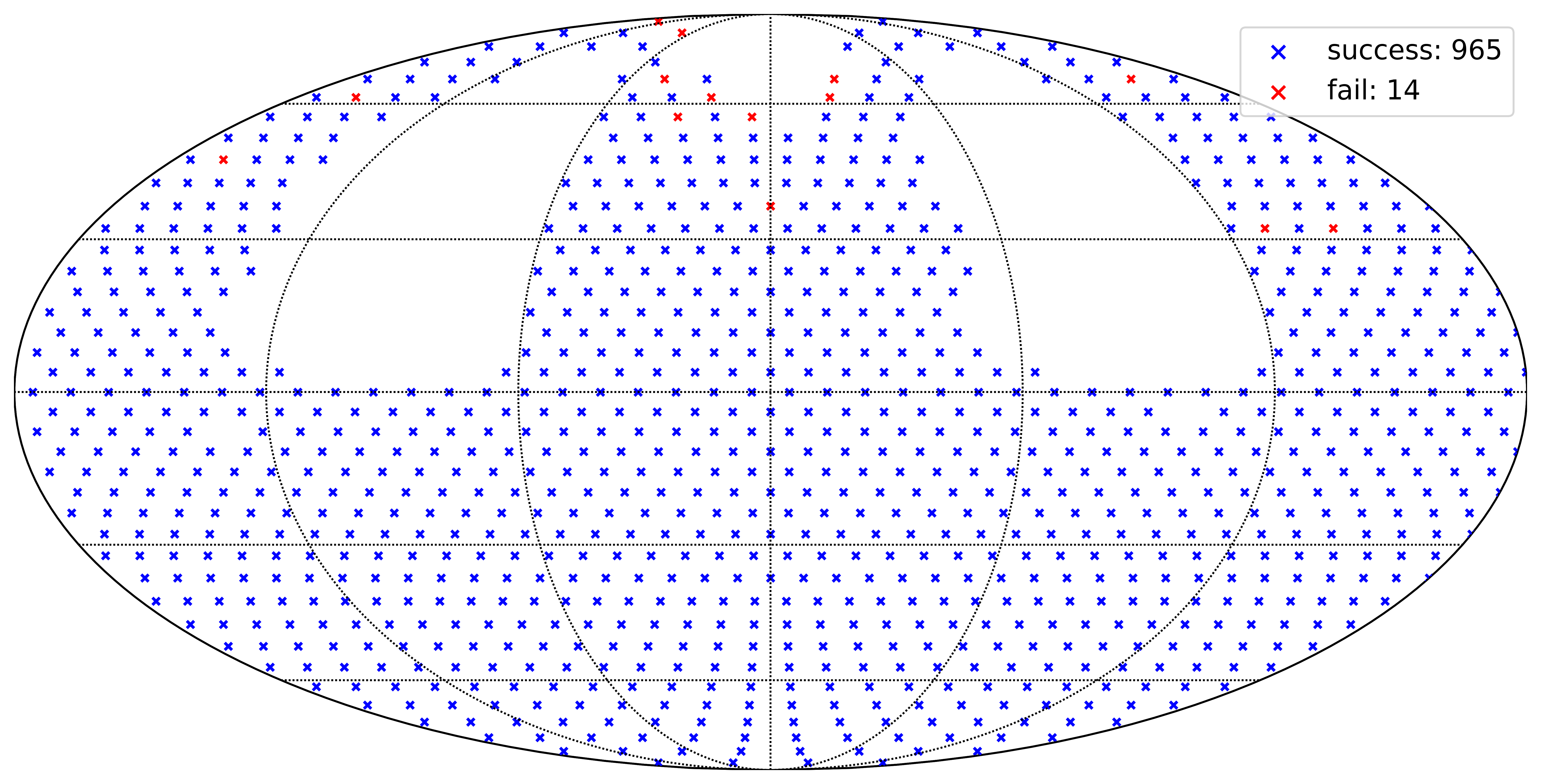}%
    \label{fig:trajectory-2d-curved-spherical-direct}
  }

  \caption{
    \textbf{Reachability}.
    Reachability maps are plotted for each trajectory type (circular vs. spherical, columns) and each gripper type (straight vs. curved, rows).
    The plots resemble a 2D cartographic projection (Mollweide projection) of the 3d sphere surface.
    The circular trajectories cover a fraction of the sphere surface, representing all possible sample rotations.
    The robotic arm can reach all points on the circle (left column).
    For the spherical trajectory, the robotic arm cannot reach all way-points, hence the blind spots on the maps (right column).
    The red points indicate way-points with successful motion planning but failure during execution (see section \ref{sec:path_planning} for details).
    We achieved the best coverage of the sphere (81.6 \%) with a curved gripper for the sample holder (lower right).
  }
  \label{fig:reachability}
\end{figure*}

In Fig. \ref{fig:reachability}, we plotted each experiment's reachability as a two-dimensional coverage map.
The trajectory way-points represent rotations of the sample as 3d points on the sphere surface.
We projected these points to 2d with the cartographic Mollweide projection \cite[pp. 112-113]{snyder1997flattening} for improved visualization of the coverage of the surface area.

We differentiate the two trajectory types, circular and spherical.
The maps visualize coverage maps on the rows for the two gripper types (straight and curved).
While the interior of the projected sphere surface includes all possible rotations in 3d and hence all rotations of the sample, only those rotations where the robotic arm has a valid and collision-free configuration are plotted with cross markers.
The empty (white) spots resemble rotations the arm cannot reach.
It is important to note that the circular trajectory only attempted a fraction of the rotations on the sphere surface because of the trajectory type.

\begin{figure*}[t]
  \centering
  \subfloat[Spherical trajectory, curved gripper - no obstacles]{\includegraphics[width=.45\textwidth]{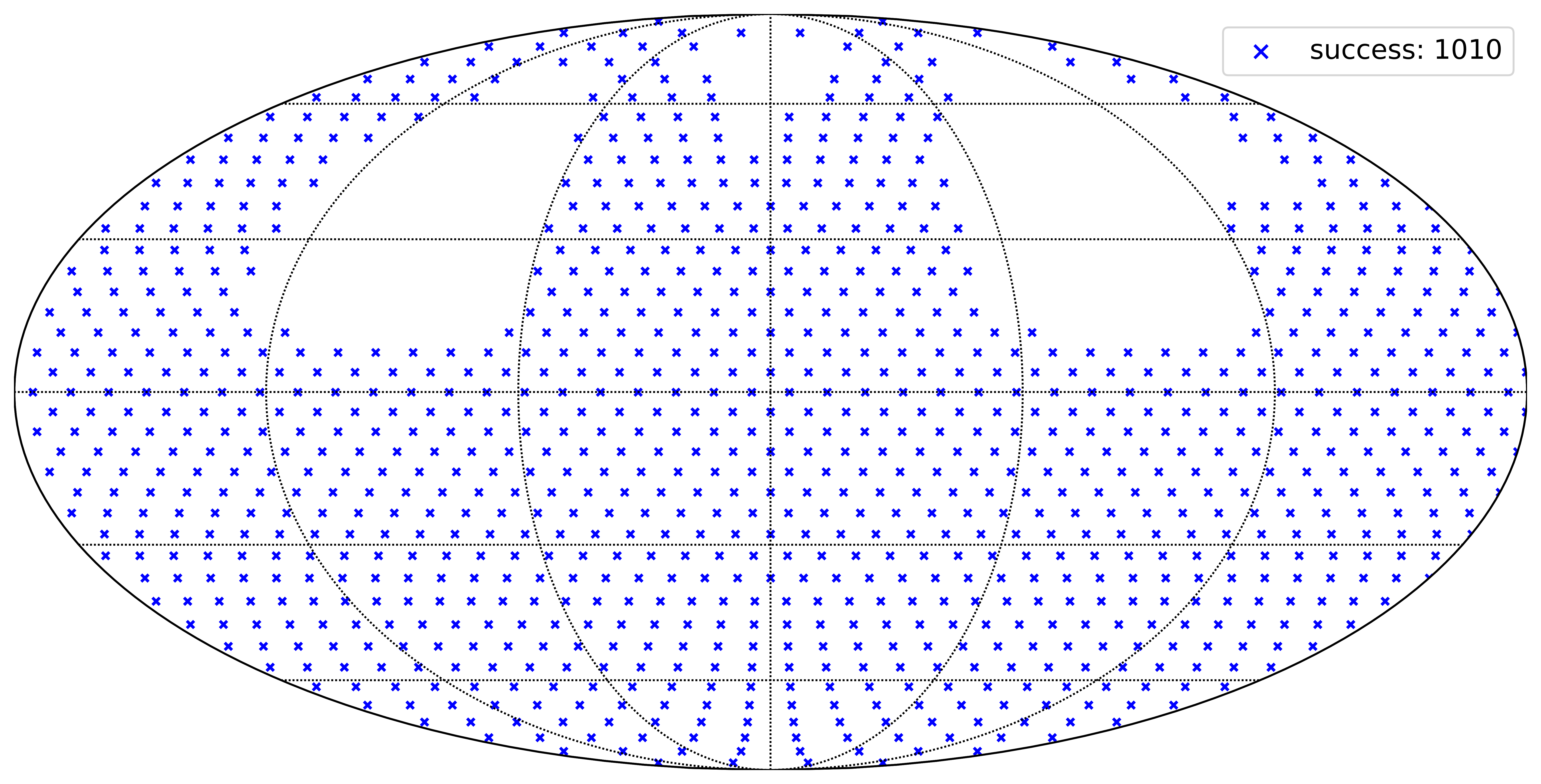}%
    \label{fig:trajectory-2d-direct-only-table}
  }
  \hfil
  \subfloat[Spherical trajectory, curved gripper - occlusions of sample by arm allowed]{\includegraphics[width=.45\textwidth]{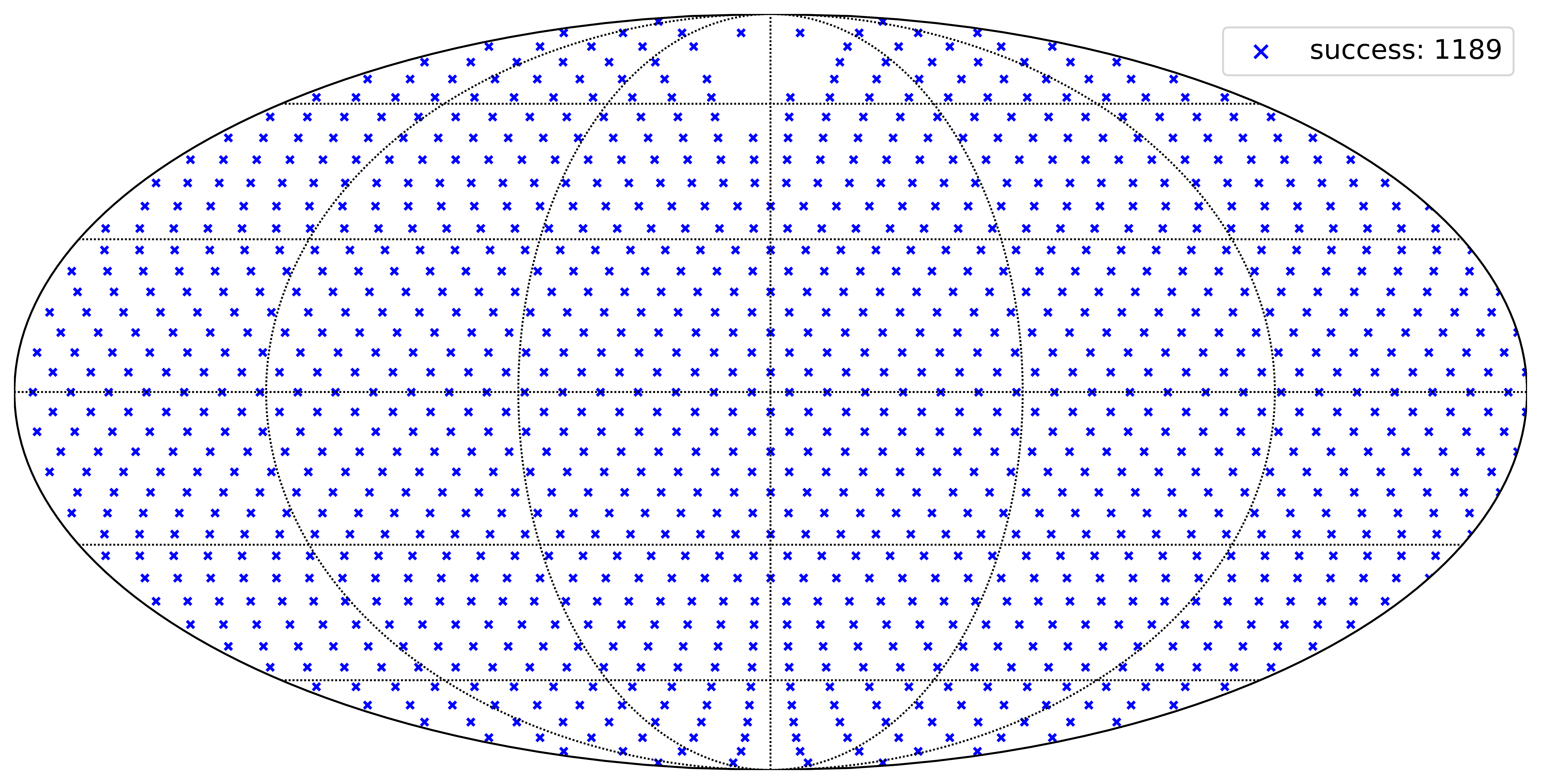}%
    \label{fig:trajectory-2d-direct-only-table-no-xray}
  }

  \caption{
    \textbf{Theoretical Reachability}.
    We plotted theoretical reachability maps for spherical trajectories for two situations only possible in the simulation.
    On the left, we removed all obstacles surrounding the robotic arm in the laboratory setup from the configuration space except for the table.
    On the right, in addition to removing all collision objects, we removed the x-ray beam encoded as a cubic collision object between the source and the detector.
    This collision object prevents occlusions of the sample by the robotic arm on the detector images.
    On the left, we achieved a successful planning rate of 84.2\% (1010/1200), and on the right, 99.1\% (1189/1200).
    The reachability number 99.1\% proves the flexibility of the given robotic arm in combination with the curved gripper.
  }
  \label{fig:reachability-theoretical}
\end{figure*}

We have also plotted two additional reachability maps where we simulated two scenarios only possible in the simulation (see Fig. \ref{fig:reachability-theoretical}).
For the first plot, we removed all obstacles from the configuration space except for the table, which should represent the ideal situation where a hutch is constructed specifically for the robotic sample holder.
For the second plot, we also removed the X-ray beam collision object from the configuration space to calculate the highest possible coverage for the robotic arm.
The X-ray beam is modeled as a collision because it avoids occlusions of the arm with the sample on detector images by invalidating such configurations of the robotic arm.

\subsection{CT measurements}\label{sec:experiments-measurements}
We conducted four experiments: one sample was measured with two different gripper parts (straight and curved) for the sample holder on two trajectories (circular and spherical).

\begin{figure*}[t]
  \centering
  \subfloat[Sample side]{\includegraphics[width=.33\textwidth]{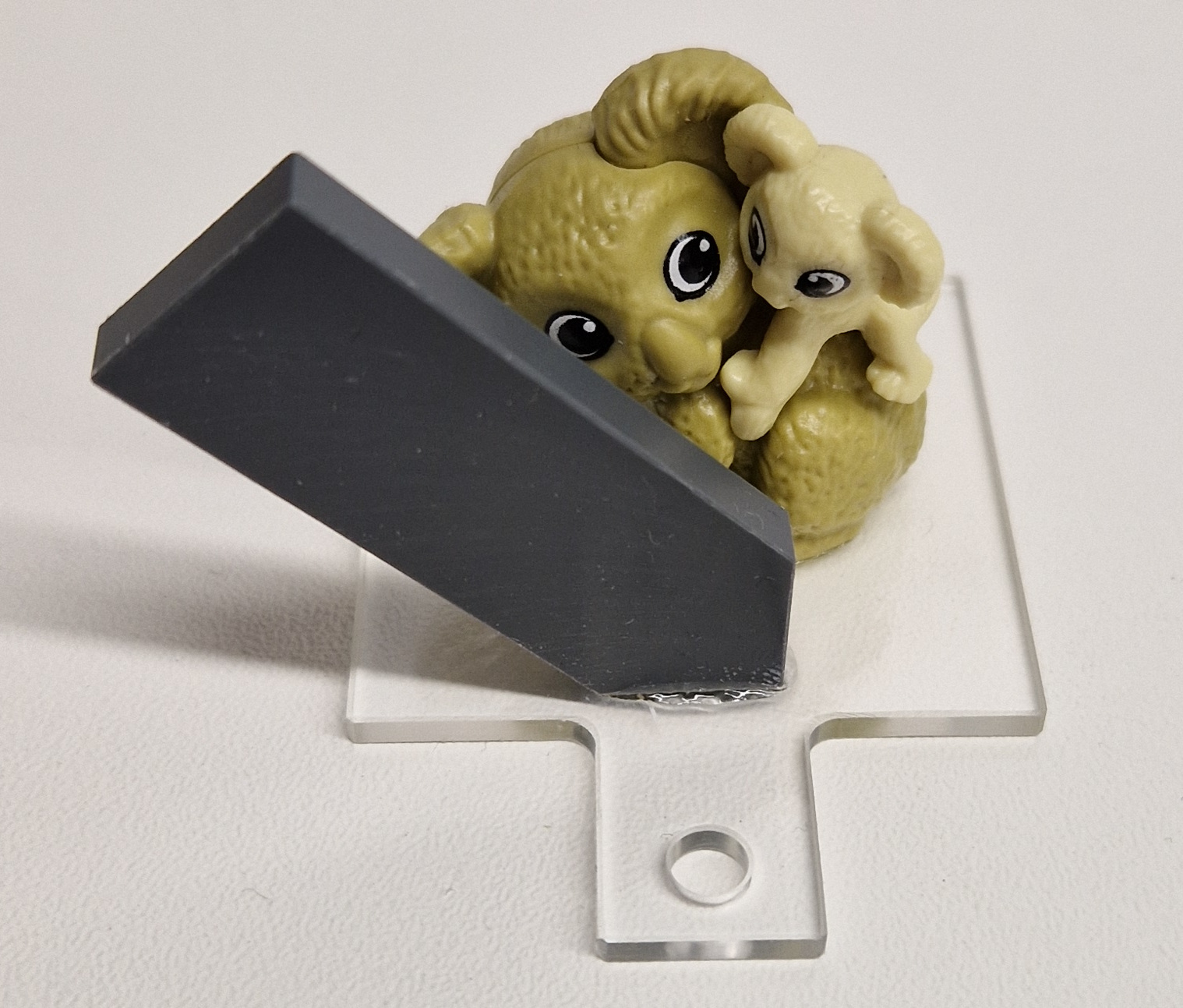}%
    \label{fig:sample-1}
  }
  \hfil
  \subfloat[Sample front]{\includegraphics[width=.33\textwidth]{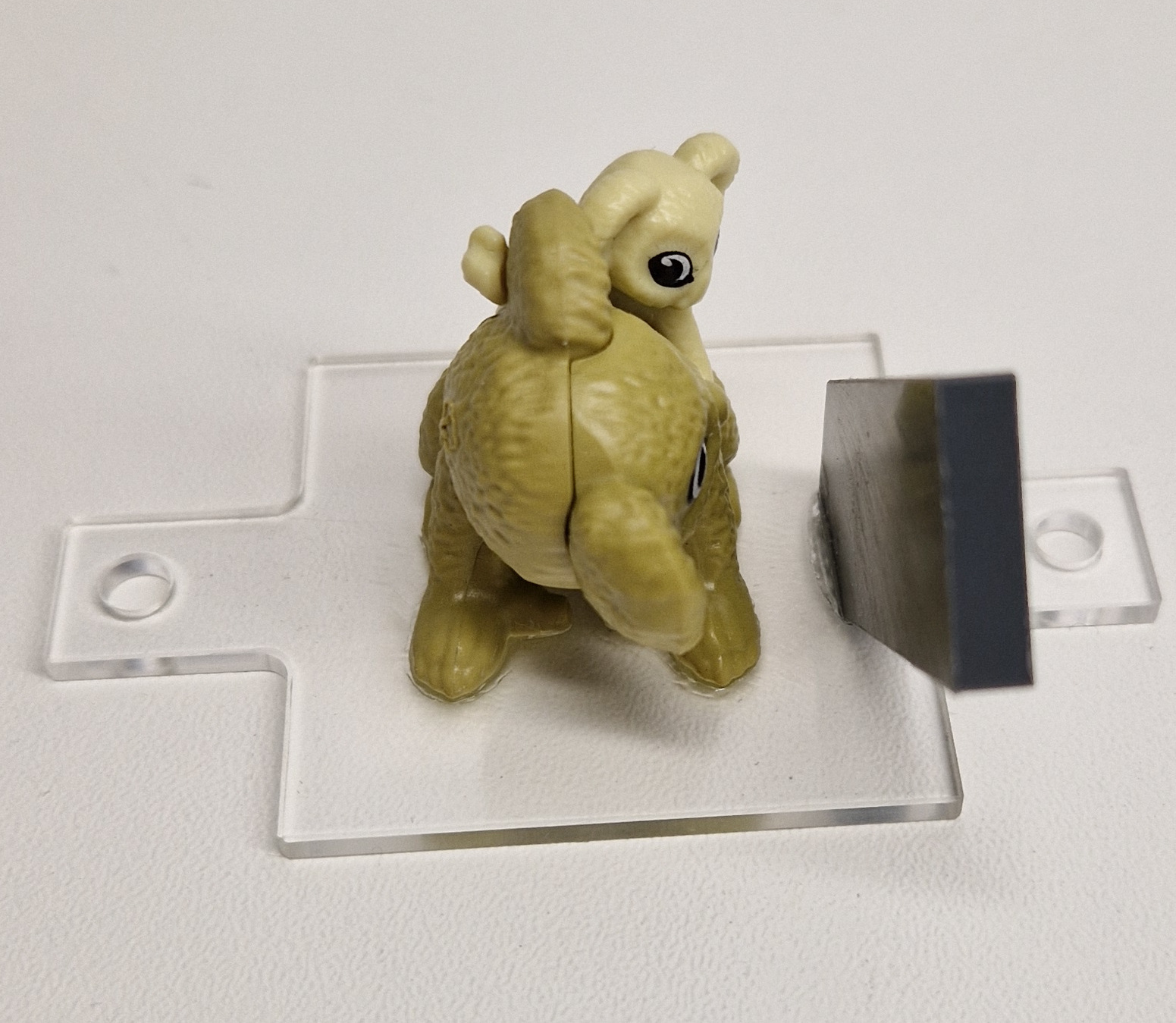}%
    \label{fig:sample-2}
  }

  \caption{
    \textbf{Sample}.
    The sample consists of the object of interest (a toy brick) and an absorber (polyvinyl chloride plate), which were manually glued to a Plexiglass mounting plate.
    The toy brick has dimensions 31 x 21 x 31 mm, and the absorber has a thickness of 4 mm.
    The absorber plate has a significantly higher X-ray contrast absorption rate than the toy brick.
    We aim to introduce beam hardening artifacts with this property in the reconstructions and evaluate the performance of different trajectory types in tackling this issue.
  }
  \label{fig:sample}
\end{figure*}

The sample we used for all experiments is displayed from two perspectives in Fig. \ref{fig:sample}.
It consists of two separate parts: a bunny toy brick (Fig. \ref{fig:sample-2}, left) and a solid piece of polyvinyl chloride (PVC) with a thickness of 4 mm (Fig. \ref{fig:sample-2}, right).
We chose this composition because these two parts differ significantly in their absorption rate, which helps compare circular and spherical trajectories in their reconstruction performance for image quality.
We have cut the absorber plate in a non-orthogonal shape relative to the mounting plate and arranged it next to the toy brick to cause beam-hardening artifacts in the reconstructions of our experiments with this sample.
One can execute the circular trajectory with conventional methods like a rotational stage, but a flexible sample holder like ours is necessary to execute the spherical trajectory.
We aim to demonstrate the superiority of spherical trajectories for complex samples.

\begin{figure*}[h]
  \centering
  \subfloat[Reconstruction slices]{\includegraphics[width=0.8\textwidth]{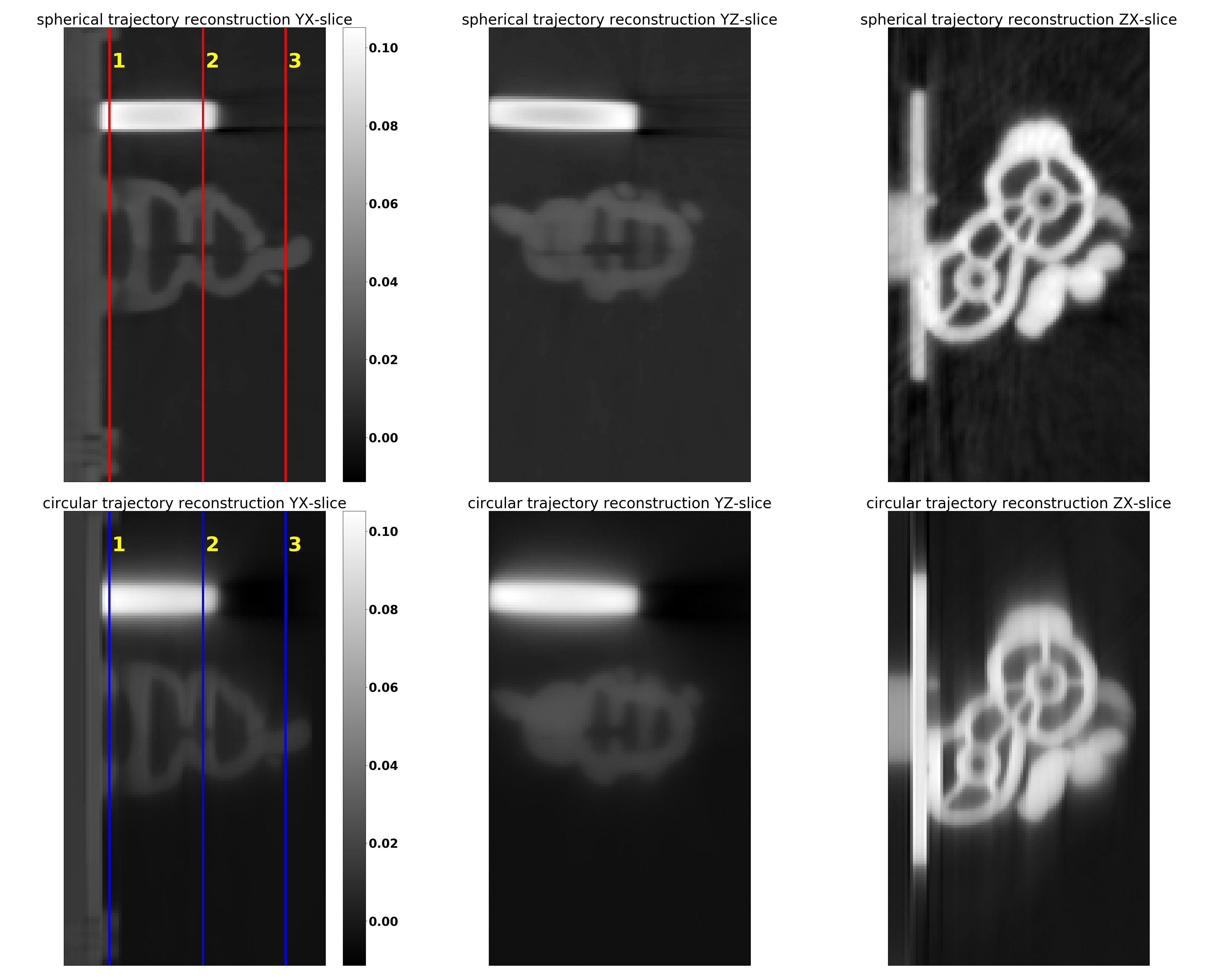}%
    \label{fig:reconstructions-slices}
  }
  \hfil
  \subfloat[Reconstruction line profiles]{\includegraphics[width=\textwidth]{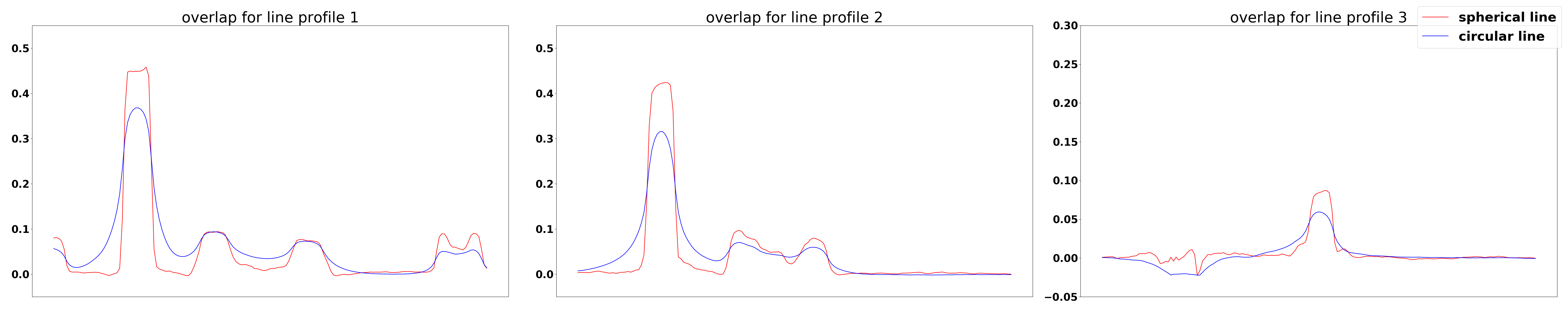}%
    \label{fig:reconstructions-lin-profiles}
  }

  \caption{
    \textbf{Experimental Results}.
    A sample was measured and reconstructed with the robotic arm with the straight gripper part to compare the circular trajectory (conventional) with the spherical trajectory (advanced).
    The reconstruction volumes are registered and aligned with our calibration algorithm (see section \ref{sec:calibration}).
    We binned the detector images with $4*4$, and the reconstruction volume has dimensions $720^{3}$.
    % The selected slices are from the perspective of the x-ray source.
    A zoom factor of 5x was applied to the slices to crop the region of interest.
    We plotted line profiles at three different positions for the YX slices.
    Our observation is that the reconstruction of the measurements with the spherical trajectory (top left, red lines) is superior qualitatively and quantitatively compared to the reconstruction of the circular trajectory (bottom left, blue lines).
    Qualitatively, there are fewer artifacts, and the image is sharper than with the circular trajectory.
    From a quantitative perspective, the line profiles for the spherical trajectory (red) are steeper than those for the circular trajectory (blue), making the image sharper.
  }
  \label{fig:reconstructions}
\end{figure*}

For each CT measurement, we acquired the images with a source voltage of $45$ kV, source power of $1445 \mu$A, and exposure time of 1s.
In Fig. \ref{fig:reconstructions-slices}, the reconstruction of our sample is shown from three different perspectives (YX, YZ, and ZX) for the two trajectory types and the straight gripper part.
All volumes are registered with each other due to the calibration process\cite{Pekel_2022}, as the center of the helix structure serves as the coordinate system's origin.
We plotted line profiles in Fig. \ref{fig:reconstructions-slices} for three different cross-sections of the central YX slice of the sample.

\section{Discussion}\label{sec:discussion}

In this section, we will discuss the results of our experiments from section \ref{sec:experiments-and-results}, where we aimed to measure our system's performance for trajectory reachability and reconstruction image quality.
We categorized the experiments by trajectory and gripper part type.
The trajectory type affects the reconstruction image quality and completeness.
% Our analysis is based on conventional circular trajectories and more advanced spherical trajectories.
The choice of gripper part type affects the system's ability to reach specific goals and the probability of triggering collision reflexes of the robotic arm's control box.
% In section \ref{sec:experiments-sample-holder-gripper-types}, we have stated that we introduced the two gripper part types: straight and curved.

\subsection{Reachability analysis}\label{sec:discussion-reachability-analysis}
In Fig. \ref{fig:reachability}, we plotted each experiment's reachability as a two-dimensional coverage map.
Additionally, we have listed the absolute and relative reachability and trajectory execution success numbers in tables \ref{tab:reachability-statistics} and \ref{tab:trajectory-execution-statistics}.

\subsubsection{Sample holder gripper type}\label{sec:discussion-sample-holder-gripper-types}
Choosing different types of grippers results in different unreachable regions of the sphere.
For both gripper types, we have an explanation for the two blind spots near the equator:
We have modeled the X-ray beam as a cubic collision object in the configuration space of the robotic arm in our motion planning pipeline.
This modeling means that unless contact of a specific link with the X-ray beam is explicitly allowed as an exception, all configurations of the arm that place one of the links into the X-ray beam are marked as a potential collision and hence as invalid configurations.
At these rotations, one of the links would cover the sample on the detector image if an image was acquired.

Two additional issues arise more often with the straight sample holder gripper.
The first one is that collision reflexes are triggered more often when approaching spots on the bottom of the sphere because the last link of the arm has to approach the way-point at a spot close to the first link.
With the curved gripper, we can increase the distance between the first and last link and limit this behavior.
The second issue is that way-points that lie at the opposite end of the arm's mounting position are harder to reach for the arm with increasing sample size as the maximum reach of the arm is physically limited by the link lengths.
The dependence on the sample size occurs since we center the sample at the central X-ray, which puts the last link of the arm further away from the central X-ray.

\subsubsection{Trajectory type}\label{sec:discussion-trajectory-types}
We can see that the robotic arm has no difficulties reaching every single point on the circle with the straight gripper part for the circular trajectory.
The circle is parallel to the table where the arm is also mounted.
With the curved gripper, the arm cannot reach a fraction of the circular trajectory (7.2\%, see table \ref{tab:reachability-statistics}) as the links collide with the X-ray beam collision object (see section \ref{sec:discussion-sample-holder-gripper-types}) which would result in occlusions on the detector images.
In contrast, we can see that the robotic arm cannot reach way-points in some areas of the sphere for the spherical trajectory, meaning that the reconstruction will lack images from specific rotations of the sample.
The unreachable region's location and size depend on the gripper's choice.
The potential coverage rate lies between 75.6\% and 81.6\% out of 1200 potential way-points.

Furthermore, we can conclude from the numbers in tables \ref{tab:reachability-statistics} and \ref{tab:trajectory-execution-statistics} that with the right choice of sample holder gripper type, the actual coverage of the sphere surface can be further increased when considering issues that arise during trajectory execution (see section \ref{sec:path_planning}).
The coverage rates of 75.6\% and 81.6\% mentioned above decrease further while the robotic arm executes the planned trajectory: The arm only reaches 95.4\% and 98.6\% of the given coverage rates due to collision reflexes of the robotic arm explained in section \ref{sec:path_planning}.
However, we can see that the gripper type affects the error rate during execution significantly, as with the curved gripper, the robotic arm only misses out 1.4\% of the trajectory.

In Fig. \ref{fig:reachability-theoretical}, we have plotted theoretical reachability maps that are only possible in simulation.
We can conclude from Fig. \ref{fig:trajectory-2d-direct-only-table} that the reachability could be increased from 81.6\% to 84.2\% by installing the robotic arm in an extensive safety hutch with more space inside.
Furthermore, Fig. \ref{fig:trajectory-2d-direct-only-table-no-xray} shows that the given robotic arm can reach 99.1\% of the sphere surface with the curved gripper type if we tolerate occlusions of the sample by the links on the detector images.
In our experiments, we do not, because occlusions with the cylindrical part of the sample holder would prevent successful calibration and occlusions with the sample would cause artifacts in the reconstruction.

\subsection{CT measurements}\label{sec:discussion-measurements}
From a qualitative and quantitative perspective, we can discuss our system's performance by examining the results in Figure \ref{fig:reconstructions}.

Qualitatively, we can see in Fig. \ref{fig:reconstructions-slices} that the slices depicted in the top row (spherical trajectory, straight gripper) are sharper overall when compared to the slices in the bottom row (circular trajectory, straight gripper).
When examining the region between the absorber at the top and the toy brick in the middle (YX-slices), we can see that the slice of the experiment with the spherical trajectory does not cause artifacts. Hence this area is genuinely black compared to the slice on the bottom, where we can spot white traces.
We can also spot significant differences for the slices in the center and right columns of Fig. \ref{fig:reconstructions-slices}, for example, on the right column, the inner structure of the toy brick is much sharper for the spherical trajectory (top right) when compared to the circular trajectory (lower right).

For a more quantitative comparison of the reconstructions, we have plotted line profiles at three different locations of the YX-slices in Fig. \ref{fig:reconstructions-lin-profiles}.
The first line profile crosses the absorber, the toy brick, and one of the screws used for mounting the sample plate to the cylinder.
The second line crosses the absorber and the toy brick, and the third line only crosses the toy brick.

For line profiles one and two, we can see that the red line (spherical trajectory) has a steeper curve and hence a higher derivative for the absorber compared to the blue line (circular trajectory), which means that the image is sharper with the spherical trajectory.
Additionally, we can see that at the end of the first line profile plot, the red line can reflect the screw head in contrast to the blue line, where the screw head does not have a notch.
We verified the visual increase in sharpness by measuring the gradient magnitude of the line profiles in Fig. \ref{fig:reconstructions-lin-profiles}.
The gradient magnitudes of the line profiles 1,2 and 3 improved by 50\%, 58\% and 80\% respectively with the spherical reconstruction compared to the circular.

Another critical observation is that the absorber causes artifacts with two different trajectory types in two orthogonal directions.
In the case of the circular trajectory, the artifacts are parallel to X-ray beams.
The artifacts are parallel to the absorber and orthogonal to the X-ray beams for the spherical trajectory.

Furthermore, in table \ref{tab:calibration-statistics}, we can see that the calibration success numbers are between 98.0\% and 99.8\% for all experiments.
Achieving a high success rate in the calibration is crucial, as only those images with a successful calibration are usable in the reconstruction step. Hence, the calibration influences the reconstruction image quality and completeness.
In this case, we can see that the calibration success rates are close to 100\%, and in absolute numbers, we missed the highest number of images in the case of the spherical trajectory and the curved gripper part with 19 out of 965 images.
10 out of 19 images capture poses that place the cylindrical part of the sample holder part containing the helix structure parallel to the X-ray beam, where calibration becomes very difficult because the spheres overlap with each other on the image and cannot be segmented.
The curved gripper makes these poses easier to reach as they would otherwise cause occlusions by placing the arm's links into the picture.
Our motion planning pipeline avoids these poses, especially for the straight gripper, by modeling the X-ray beam as a cubic collision object (see section \ref{sec:path_planning}).
Hence the inferior calibration statistics with the curved gripper.

\subsection{Future work}\label{sec:discussion-future-work}
We plan to improve the proposed system in the future in several ways.

We plan to design new gripper types that further maximize the coverage of the sphere surface.
We could also improve the cylinder part of the sample holder with the mounting plate for mounting bigger and heavier samples.
Currently, we glue the sample to the mounting plate, which could cause issues for heavier samples.

Moreover, experiments with optimized trajectories are the subject of future work.
We expect the system to fully benefit from the flexibility of the robotic sample holder once it can determine highly-absorbing, problematic parts of the spherical trajectory.
The authors used task-specific trajectories successfully in \cite{fischer2016object} to reduce artifacts in X-ray CT.
We plan to extend our system with task-specific acquisition strategies to fully benefit from its ability to manipulate the sample arbitrarily.

\section{Conclusion}
In this work, we have demonstrated using a seven DoF robot as a sample holder for acquiring spherical X-ray computed tomography trajectories using our unified software package with path planning, collision detection, and calibration.
Initially, we stated that the image quality of 3d reconstructions would benefit significantly from spherical trajectories when the sample contains highly absorbing parts.
Our findings have confirmed that with a spherical trajectory, the image quality is superior qualitatively and quantitatively compared to a conventional circular trajectory.

% \section*{Acknowledgments}

\bibliography{references.bib}{}
\bibliographystyle{IEEEtran}

\end{document}